\documentclass[twoside]{article}

\usepackage[accepted]{aistats2020}
\usepackage{times}
\usepackage{amsmath}
\usepackage{amssymb}
\usepackage{amsopn}
\usepackage{epsfig}
\usepackage{microtype}
\usepackage{subfigure}
\usepackage{booktabs}
\usepackage{url}
\usepackage{booktabs}
\usepackage{amsfonts}
\usepackage{nicefrac}
\usepackage{microtype}
\usepackage{algorithm}
\usepackage{algorithmic}
\usepackage{wrapfig}
\usepackage{xcolor}

\newtheorem{th_innerprod_kern}{Theorem}

\newtheorem{th_kernel_weight}[th_innerprod_kern]{Theorem}

\newtheorem{def_2qkernel}{Definition}
\newtheorem{def_freekernel}[def_2qkernel]{Definition}

\newcommand{\tsp}[0]{{\rm T}}
\newcommand{\llp}[0]{\left(}
\newcommand{\ggp}[0]{\right)}

\newcommand{\llpt}[0]{(}
\newcommand{\ggpt}[0]{)}
\newcommand{\epow}[0]{\odot}

\newcommand{\infset}[1]{{\mathbb{#1}}}
\newcommand{\finset}[1]{{\tt #1}}
\newcommand{\distrib}[1]{{\mathcal{#1}}}

\newcommand{\normdist}[0]{\distrib{N}}

\newcommand{\gp}[0]{\mathcal{GP}}

\newcommand{\citep}[1]{\cite{#1}}

\DeclareMathOperator\sgn{sgn}

\DeclareMathOperator\argmax{argmax} 

\definecolor{Gray}{gray}{0.9}

\begin{document}

\twocolumn[

\aistatstitle{Accelerated Bayesian Optimization through Weight-Prior Tuning}

\aistatsauthor{$\begin{array}{c} \mbox{Alistair Shilton}^1 \\ \mbox{Sunil Gupta}^1 \\ \mbox{Santu Rana}^1 \\ \end{array}\!\!\!\!\!\!\!\!$ 
\And $\begin{array}{c} \mbox{Pratibha Vellanki}^2 \\ \mbox{Laurence Park}^3 \\ \end{array}\!\!\!\!\!\!\!\!$ 
\And $\begin{array}{c} \mbox{Cheng Li}^4 \\ \mbox{Svetha Venkatesh}^1 \\ \mbox{Thomas Dorin}^5 \\ \end{array}\!\!\!\!\!\!\!\!$
\And $\begin{array}{c} \mbox{Alessandra Sutti}^5 \\ \mbox{David Rubin}^5 \\ \end{array}\!\!\!\!\!\!\!\!$
\And $\begin{array}{c} \mbox{Teo Slezak}^5 \\ \mbox{Alireza Vahid}^5 \\ \mbox{Murray Height}^5 \\ \end{array}$}

\aistatsaddress{$\begin{array}{c} \\
\!{}^1\mbox{Applied Artificial Intelligence Institute (${\rm A}^2{\rm I}^2$), Deakin University, Geelong, Australia} \\ 
\!{}^2\mbox{Office of National Statistics, Southampton, United Kingdom} \\ 
\!{}^3\mbox{School of Computing, Engineering and Mathematics, University of Western Sydney, Sydney, Australia} \\ 
\!{}^4\mbox{School of Computing, National University of Singapore, Singapore} \\ 
\!{}^5\mbox{Institute for Frontier Materials (IFM), Deakin University, Geelong, Australia} \\ 
\end{array}$ } ]





\begin{abstract}
\vspace{-0.1cm}
Bayesian optimization (BO) is a widely-used method for optimizing expensive (to 
evaluate) problems.  At the core of most BO methods is the modeling of the 
objective function using a Gaussian Process (GP) whose covariance is selected 
from a set of standard covariance functions.  From a weight-space view, this 
models the objective as a linear function in a feature space implied by the 
given covariance $K$, with an arbitrary Gaussian weight prior ${\bf w} \sim 
\normdist ({\bf 0},{\bf I})$.  In many practical applications there is data 
available that has a similar (covariance) structure to the objective, but 
which, having different form, cannot be used directly in standard transfer 
learning.  In this paper we show how such auxiliary data may be used to 
construct a GP covariance corresponding to a more appropriate weight prior for 
the objective function.  Building on this, we show that we may accelerate BO by 
modeling the objective function using this (learned) weight prior, which 
we demonstrate on both test functions and a practical application to 
short-polymer fibre manufacture.
\end{abstract}

\vspace{-0.2cm}
\section{Introduction}
\vspace{-0.2cm}

Bayesian Optimization (BO) \citep{Sno1,Bro2} is a form of sequential 
model-based optimization (SMBO) that aims to to find ${\bf x}^* = \argmax_{{\bf 
x}} f ({\bf x})$ with the least number of evaluations for an expensive (to 
evaluate) function $f : \infset{X} \to \infset{R}$.  In many cases there exists 
an additional, auxiliary dataset $\finset{A}$ relating to $f$ that is relevant 
to the problem at hand - for example patents and technical handbooks that 
contain condensed knowledge related to $f$, data obtained from similar 
(possibly older or superseded) equipment and, more generally, results from 
optimizing structurally similar functions.  It is not necessary that the 
auxiliary data be generated by $f$ (if $\finset{A}$ was generated by a 
function sufficiently similar to $f$ then we could simply use standard transfer 
learning techniques \citep{Pan2,Yog1,Bar5,Joy1,Shi21}); rather, what matters is 
that the auxiliary data is structurally similar to $f$ insofar as it has a 
similar {\em covariance} structure.  Human experimenters routinely use such 
data to inform and accelerate the (experimental) optimization process.  In the 
present paper we investigate the use of such data to accelerate Bayesian 
Optimization.

Typically in BO it is assumed that $f \sim \gp (0,\tilde{K})$ is a draw from a 
zero mean {G}aussian {P}rocess (GP) \citep{Ras1} with covariance $\tilde{K} : 
\infset{X} \times \infset{X} \to \infset{R}$.  In general $\tilde{K}$ is 
unknown, so we select an alternative $K$ to use in its place.  When selecting 
$K$ we use intuitions and experimentalist knowledge relating to $f$ - for 
example how smooth is $f$ (should we use squared exponential or Mat{\'e}rn 
covariance?), is the length-scale consistent everywhere (should be use an 
isotropic or anisotropic covariance?) etc, but typically we restrict our choice 
to a subset of standard kernels.  This introduces approximations and experimenter 
biases, as our knowledge of the covariance structure of $f$ is by definition 
incomplete in most cases, and moreover it appears unlikely that the covariance 
structure can captured by a small set of generic ``standard'' covariance 
functions.  Subsequently BO is likely to converge more slowly than it should as 
the GP used to model $f$ has an inaccurate covariance function.

To overcome the deficiencies it is standard practice to tune hyper-parameters of 
$K$ at each iteration of the BO, either to maximize the log-likelihood function 
or minimize some sort of approximation error (e.g. leave-one-out error).  The 
auxiliary dataset may be used to accelerate the process by pre-tuning 
hyper-parameters or, if it is large enough, techniques such as multi-kernel 
learning (MKL) \citep{Lan2,Bac1} may be used to find to find a more nuanced 
fit.  Nevertheless there remains an underlying assumption that the space 
spanned by the chosen subset of standard kernels contains a good approximation 
of the actual covariance.

In this paper we provide a principled alternative approach to covariance 
construction using auxiliary data.  Starting with the weight-space perspective 
of GPs \cite{Ras2}, our approach pre-selects a weight prior by modeling the 
auxiliary data using kernel methods and then transfers this to the BO problem 
using $m$-kernel theory.  From the weight-space perspective a GP models $f \sim 
\gp (0,K)$ as a linear function in feature space $f ({\bf x}) = {\bf w}^{\tsp} 
{\mbox{\boldmath $\varphi$}} ({\bf x})$ with a weight prior ${\bf w} \sim 
\normdist ({\bf 0}, {\bf I})$ \cite{Ras2}, where the feature map 
${\mbox{\boldmath $\varphi$}} : \infset{X} \to \infset{R}^d$ is implicitly 
defined by the covariance (kernel) $K$ via Mercer's condition \cite{Mer1}.  
Thus when we use a standard covariance $K$ we are stating a belief that:
\begin{enumerate}
 \item \vspace{-0.2cm} $f$ is linear in the weight space defined by $K$. 
 \item \vspace{-0.2cm} The features $\varphi_i ({\bf x})$ are uncorrelated and equally 
       important in this approximation (hence ${\bf w} \sim \normdist ({\bf 0}, 
       {\bf I})$).
\end{enumerate}
For universal kernels \citep{Mic1,Sri2} such as the squared-exponential (SE) 
kernel or the Mat{\'e}rn kernels the first assumption will be true for most 
(reasonably well-behaved) functions $f$, as such kernels can approximate most 
(reasonably well-behaved) functions $f$ to arbitrary accuracy.  By contrast, 
the second assumption is much more tenuous.  It will almost certainly be 
incorrect if the kernel hyper-parameters are incorrectly specified, and there 
is no a-priori reason to believe it will be accurate in general.  This 
motivates us to replace the weight prior with the more general weight prior 
${\bf w} \sim \normdist ({\bf 0}, {\mbox{\boldmath $\Sigma$}}_{\mbox{\boldmath 
$\tau$}})$, where the covariance ${\mbox{\boldmath $\Sigma$}}_{\mbox{\boldmath 
$\tau$}}$ is a diagonal matrix learned from the auxiliary data.  This embodies 
the alternative belief:\vspace{-0.1cm}
\begin{description}
 \item [\normalfont2 (alt).] \vspace{-0.2cm} The features $\varphi_i ({\bf x})$ 
       are uncorrelated {\em and their relative importance may inferred from 
       the auxiliary dataset $\finset{A}$} (hence ${\bf w} \sim \normdist 
       ({\bf 0}, {\mbox{\boldmath $\Sigma$}}_{\mbox{\boldmath $\tau$}})$).
\end{description}

\vspace{-0.1cm} 
To achieve our goal requires two steps, namely (1) a heuristic to extract 
relative feature importance information from the auxiliary dataset 
$\finset{A}$, and (2) a way to alter the weight-prior of a GP to reflect the 
relative feature importance information found in step (1).

To address the first goal we we use standard machine (kernel) learning 
techniques.  If we apply a support vector machine (SVM) \citep{Cor1} method (or 
similar) equipped with kernel $K$ to the auxiliary dataset $\finset{A}$ then 
the answer we obtain takes the form of the representation $\alpha_0, \alpha_1, 
\ldots \in \infset{R}$ of a weight vector $\tilde{\bf w} = \sum_i \alpha_i 
{\mbox{\boldmath $\varphi$}} ({\bf x}_i)$ in feature space.  Assuming a 
reasonable ``fit'' we hypothesize that the weights $\tilde{w}_i$ will be larger 
in magnitude for relevant features and smaller for irrelevant ones.  Thus we 
generate (implicit) information about feature relevance for $\finset{A}$.

To address the second goal we borrow from $\ell^p$-norm regularization 
\citep{Sal2,Sal1} and large-margin $L^p$-moment classifiers\footnote{And more 
generally reproducing kernel Banach space (RKBS) theory \citep{Zha11,Fas1}.} 
\citep{Der1}, and particularly $m$-kernels (aka tensor kernels \citep{Sal1} or 
moment functions \citep{Der1}), which are an extension of kernels to 
multi-linear products in feature space.  We introduce a class of $m$-kernel 
families, the free kernels, that are expandable as weighted multi-linear 
products in feature space - that is, families of functions $K : \infset{X}^m 
\to \infset{R}$ for which there exists ${\mbox{\boldmath $\vartheta$}} : 
\infset{X} \to \infset{R}^d$, ${\mbox{\boldmath $\tau$}} \in \infset{R}^d$ such 
that:
\[
 \begin{array}{l}
  K_m \left( {\bf x}, {\bf x}', \ldots, {\bf x}'''' \right) = \sum_j \tau_j^2 \vartheta_j \left( {\bf x} \right) \vartheta_j \left( {\bf x}' \right) \ldots \vartheta_j \left( {\bf x}'''' \right)
 \end{array}
\]
where we note that many free kernels share the same unweighted feature map 
${\mbox{\boldmath $\vartheta$}}$.  By reformulating Gaussian Processes in terms 
of free kernels we demonstrate that the weights ${\mbox{\boldmath $\tau$}}$ 
play the role of (diagonal) weight prior covariances ${\mbox{\boldmath 
$\Sigma$}}_{\mbox{\boldmath $\tau$}} = {\rm diag} ({\mbox{\boldmath 
$\tau$}}^{\odot 2})$.  Finally, we show how this weight ${\mbox{\boldmath 
$\tau$}}$ may be changed to match the weight extracted from the auxiliary 
dataset $\finset{A}$, resulting in a kernel (covariance) whose weight-prior is 
tuned to suit $f$ (as we assume a-priori that $\finset{A}$ and $f$ share the 
same covariance structure).

Having derived a covariance $K^{\finset{A}}_2$ whose corresponding weight-space 
prior has been tuned to match the objective function $f$, we may proceed with 
standard Bayesian optimization modeling $f \sim \gp (0, K^{\finset{A}}_2)$, 
where we posit that the more accurate matching of $K^{\finset{A}}_2$ to the 
covariance structure of $f$ will accelerate the optimization process.  In 
section \ref{sec:apps} we demonstrate the efficacy of our algorithm in two  
applications, specifically (1) a test function with auxiliary data drawn from 
a flipped source and (2) new short polymer fibre design using micro-fluid 
devices (where auxiliary data is generated by an older, distinct device).

\vspace{-0.1cm}
Our main contributions are:
\begin{itemize}
 \item \vspace{-0.2cm} Definition of free-kernels and interpretation of free 
       kernel weights as weight priors for Gaussian Processes.
 \item \vspace{-0.2cm} Weight-prior tuning using an auxiliary dataset.
 \item \vspace{-0.2cm} Accelerated Bayesian optimization using tuned weight 
       priors (algorithm \ref{alg:modded_bbo}).
 \item \vspace{-0.2cm} Application of accelerated Bayesian optimization to 
       real-world scenarios (section \ref{sec:apps}).
\end{itemize}

\subsection{Notation}

We use $\infset{N} = \{ 0,1,\ldots \}$, $\bar{\infset{N}} = \infset{N} \cup \{ 
\infty \}$.  Column vectors are ${\bf a}, {\bf b}$, matrices ${\bf V}, {\bf W}$ 
(elements $a_i$, $W_{i,j}$).  ${\bf a} \odot {\bf b}$ is the element-wise 
product, ${\bf a}^{\epow b}$ the element-wise power, $| {\bf a} |$ the 
element-wise absolute, and ${\rm sum} ({\bf a}) = \sum_i a_i$.  The 
$m$-dot-product $\llpt \ldots \ggpt_m : (\infset{R}^n)^m \to \infset{R}$ is 
$\llpt {\bf a}, \ldots, {\bf a}'''' \ggpt_m  = {\rm sum} ( {\bf a} \odot \ldots 
\odot {\bf a}'''' )$ \citep{Dag1,Sal1}, so $\llpt {\bf x}, {\bf x}' \ggpt_2$ is 
the dot product.  The number of elements in a finite set $\finset{D}$ is 
denoted $|\finset{D}|$.

\vspace{-0.1cm}
\section{Problem Statement}
\vspace{-0.2cm}

In this paper we are concerned with solving the problem:
\vspace{-0.1cm}
\[
 \begin{array}{l}
  {\bf x} = {\rm argmax}_{{\bf x} \in \infset{X}} f \left( {\bf x} \right)
 \end{array}
\]
where $f$ is expensive to evaluate.  We further suppose that we have been given 
an auxiliary set of training data $\finset{A} = \{ ({\bf x}^{\finset{A}}_i, 
y^{\finset{A}}_i) \in \infset{X} \times \infset{R} | y^{\finset{A}}_2 = h(g({\bf 
x}^{\finset{A}}_i) + \nu_i), \nu_i \sim \normdist (0,\zeta^2) \}$ to accelerate 
the process, where $h$ in this expression depends on the type of data 
represented by $\finset{A}$ (e.g. if $\finset{A}$ is binary classification data 
then $h (z) = \sgn (z)$, and if $\finset{A}$ is regression data then $h (z) = 
z$).  The dataset $\finset{A}$ may include distilled knowledge from relevant 
patents (if $f$ is the for example the yield of an experiment), observations 
from an older iteration of $f$ (if $f$ is a refinement of a manufacturing 
process) or human-generated samples (experimenter intuition).

\vspace{-0.1cm}
Importantly, we do not assume $\finset{A}$ is generated by the function we wish 
to optimize (i.e. $f \ne g$ in general), but rather that $f \sim \gp (0, 
\tilde{K})$ is a draw from a (zero mean) Gaussian process with covariance 
$\tilde{K}$ and $g$ is well modeled using kernel $\tilde{K}$; and that $f$ and 
$g$ depend on similar {\em features} in the RKHS $\mathcal{H}_{\tilde{K}}$ 
(equivalently (isomorphically) features in Gaussian Hilbert space \cite{Jan1} 
induced by $\tilde{K}$).

\vspace{-0.1cm}
We use Bayesian Optimization (BO) here as $f$ is presumed expensive to 
evaluate.  Bayesian Optimization \citep{Sno1} is a form of sequential 
model-based optimization (SMBO) that aims to to find ${\bf x}^* = \argmax_{{\bf 
x}} f ({\bf x})$ with the least number of evaluations for an expensive (to 
evaluate) function $f : \infset{X} \to \infset{R}$.

\section{Background and Definitions} \label{sec:basics}
\vspace{-0.2cm}

The algorithm we present in this paper combines Bayesian optimization (BO) with 
a variant of the kernel trick known as the $m$-kernel trick to achieve better 
tuning of the covariance (kernel) based on auxiliary data.  As practitioners 
may not be familiar with the $m$-kernel trick, in this section we provide a 
quick review of the kernel trick and its $m$-kernel extension and their 
application to support vector machines (SVMs).  We also define families of 
$m$-kernels (free kernels) that have a useful property that will play a central 
role later.

\subsection{Kernels and the Kernel Trick}
\vspace{-0.2cm}

In machine learning the so-called ``kernel trick'' \citep{Cor1,Sch3,Cri4} is a 
ubiquitous way to converting any linear algorithm that may be expressed in 
terms of dot products in input space into a non-linear algorithm by simply 
replacing all dot products ${\bf x}^{\tsp} {\bf x}'$ with kernel 
evaluations $K ( {\bf x}, {\bf x}' )$, where $K : \infset{X} \times \infset{X} 
\to \infset{R}$ ($\infset{X} \ne \infset{R}^n$ in general) is a positive 
definite kernel (called Mercer kernels here to prevent later ambiguity).  By 
Mercer's condition \cite{Mer1}, corresponding to $K$ is an (implicit) feature 
map ${\mbox{\boldmath $\varphi$}} : \infset{X} \to \infset{R}^d$, $d \in 
\bar{\infset{N}}$, such that $K ({\bf x}, {\bf x}') = {\mbox{\boldmath 
$\varphi$}} ({\bf x})^{\tsp} {\mbox{\boldmath $\varphi$}}({\bf x}')$, so the 
kernel trick applies an (implicit) transform ${\bf x} \to {\mbox{\boldmath 
$\varphi$}} ({\bf x})$ to all inputs.

In Gaussian Processes (GPs) and Bayesian theory more generally the covariance 
function $K$ is analogous to the kernel function in machine learning, though it 
is often conceived differently and distinctions may arise \cite{Kan2}.  The 
performance of a kernel in a given context depends on how well matched it and 
its attendant hyper-parameters are to the problem at hand.  This matching 
problem is known as kernel selection and hyper-parameter tuning, and typically 
relies on heuristics, intuition(s)/prior(s), and optimization techniques such 
as grid search or Bayesian optimization.

In practice, the kernel $K$ is usually selected from a set of well-known 
kernels such as the linear, polynomial, squared-exponential (SE) or Mat{\'e}rn 
kernels - and potentially combinations thereof (eg. multi-kernel learning 
\citep{Lan2,Bac1}) - where any kernel parameters are simultaneously tuned to 
optimize some measure of fit such as log-likelihood or cross-validation error.  
We note that by taking this approach the search is effectively limited to the 
subset of possible feature maps embodied by the set of possible kernels chosen, 
which is by no means guaranteed to approach optimal feature map.

\vspace{-0.1cm}
\subsection{$m$-Kernels and Free Kernels} \label{sec:m-kern}
\vspace{-0.2cm}

Less well known is the $m$-kernel trick, which allows conversion of any linear 
algorithm that may be expressed in terms of $m$-dot-products 
\citep{Dag1} $\llp {\bf x}, {\bf x}', \ldots, {\bf x}'''' \ggp_m = {\rm sum} 
({\bf x} \odot {\bf x}' \odot \ldots \odot {\bf x}'''')$, $m \in \infset{N}$, 
in input space into a non-linear algorithm.  Similar to the kernel trick, it 
works by replacing $m$-dot-products $({\bf x}, {\bf x}', \ldots, {\bf 
x}'''')_m$ with $m$-kernel evaluations $K_m ( {\bf x}, {\bf x}', \ldots, {\bf 
x}'''' )$, where in this case $K_m : \infset{X}^m \to \infset{R}$ is an 
$m$-kernel (tensor kernel \citep{Sal2}) - that is, $K_m : \infset{X}^m \to 
\infset{R}$, and there exists an (implicit) feature map ${\mbox{\boldmath 
$\varphi$}}_m : \infset{X} \to \infset{R}^d$, $d \in \bar{\infset{N}}$, such 
that:
\vspace{-0.1cm}
\begin{equation}
 \begin{array}{l}
  \!\!\!\!\!\!K_m \left( {\bf x},\ldots,{\bf x}'''' \right) 
  = \llp {\mbox{\boldmath $\varphi$}}_m \left( {\bf x} \right), \ldots, {\mbox{\boldmath $\varphi$}}_m \left( {\bf x}'''' \right) \ggp_m\!\!\!\!\!\!
 \end{array}
\label{eq:2qkern}
\end{equation}
So, like the kernel trick, the $m$-kernel trick works by applying an (implicit) 
transform ${\bf x} \to {\mbox{\boldmath $\varphi$}}_m ({\bf x})$ to all inputs.  
The canonical example of the $m$-kernel trick is the $p$-norm SVM.  For 
instructive purposes we have included an introduction in the supplementary 
material.  Mercer kernels are $2$-kernels.

From a Bayesian perspective an $m$-kernel $K_m$ is analogous to a 
moment function \cite{Der1}.  The formal equivalence of the $L^m$-moment 
classifier \cite{Der1} and the $\ell^p$-SVM \cite{Sal2} may be seen by 
inspection. 
Like a Mercer 
kernel, the performance of an $m$-kernel depends on its hyper-parameters.

\begin{table}
 \centering
 \begin{tabular}{| l || l | c |}
 \hline
   & $K_m \left( {\bf x}, \ldots, {\bf x}'''' \right)$ \\
 \hline
 \hline
 $\!$Linear:                & $\!\llp {\bf x}, {\bf x}', \ldots, {\bf x}'''' \ggp_m^{{\;}^{{\;}^{\;}}}$          \\ 
 $\!$Polynomial:            & $\!\left( \llp {\bf x}, {\bf x}', \ldots, {\bf x}'''' \ggp_m + \nu \right)^p$      \\ 
 $\!$Hyperbolic sine:$\!\!$ & $\!\sinh \left( \nu \llp {\bf x}, \ldots, {\bf x}'''' \ggp_m \right)$    \\ 
 $\!$Exponential:           & $\!\exp \left( \nu \llp {\bf x}, {\bf x}', \ldots, {\bf x}'''' \ggp_m \right)$     \\ 
 $\!$Log ratio:             & $\!\prod_i \ln \left( \frac{1+ x_i x_i' \ldots x_i''''}{1- x_i x_i' \ldots x_i''''} \right)$ \\ 
 $\!$SE:                    & $\!\exp ( \frac{\nu}{2} ( 2 \llp {\bf x}, \ldots, {\bf x}'''' \ggp_m - \mathop{\sum}\limits_{\cdots} \| {\bf x}^{\ldots} \|_2^2 ) )\!$ \\
 \hline
 \end{tabular}
 \caption{Examples of (free) $m$-kernels ($\nu \geq 0$, $p \in \infset{N}$).}
 \label{tab:mkernel_eg} \label{table:kern_table}
\end{table}

\vspace{-0.1cm}
For the purposes of the present paper we define the following familes of 
$m$-kernels for which the feature map is independent, in the specified sense, 
of $m$:
\vspace{-0.1cm}
\begin{def_freekernel}[Free kernel]
 A {\em free kernel} is a family of functions $K_m : \infset{X}^m \to 
 \infset{R}$ indexed by $m = 2,4,\ldots$ for which there exists an 
 {\em unweighted} feature map ${\mbox{\boldmath $\vartheta$}} : \infset{X} \to 
 \infset{R}^d$, $d \in \bar{\infset{N}}$, and feature weights ${\mbox{\boldmath 
 $\tau$}} \in \infset{R}^d$, both independent of $m$, so:
 \vspace{-0.1cm}
 \begin{equation}
  \begin{array}{l}
   \!\!\!\!\!\!K_m ( {\bf x}, \ldots, {\bf x}'''' ) 
   = \llp {\mbox{\boldmath $\tau$}}^{\epow 2}, {\mbox{\boldmath $\vartheta$}} ( {\bf x} ), \ldots, {\mbox{\boldmath $\vartheta$}} ( {\bf x}'''' ) \ggp_{m+1}\!\!\!\!\!\!
  \end{array}
  \label{eq:freekern}
 \end{equation}
 For fixed $m$ a free kernel defines (is) an $m$-kernel:
 \vspace{-0.1cm}
 \begin{equation}
  \begin{array}{l}
   \!\!\!\!\!\!K_m ( {\bf x}, \ldots, {\bf x}'''' ) 
   = \llp {\mbox{\boldmath $\varphi$}}_m ( {\bf x} ), \ldots, {\mbox{\boldmath $\varphi$}}_m ( {\bf x}'''' ) \ggp_{m}\!\!\!\!\!\!
 \vspace{-0.1cm}
  \end{array}
  \label{eq:freekern_mkern}
 \end{equation}
 with implied feature map ${\mbox{\boldmath $\varphi$}}_m ( {\bf x} ) = 
 {\mbox{\boldmath $\tau$}}^{\epow \frac{2}{m}} \odot {\mbox{\boldmath 
 $\vartheta$}} ( {\bf x} )$.
 \label{def:def_freekernel}
\end{def_freekernel}

Like Mercer kernels, standard $m$-kernels may be built - e.g. if $\infset{X} = 
\infset{R}^n$ and $k : \infset{R} \to \infset{R}$ is expandable as a Taylor 
series $k (\chi) = \sum_i \xi_i \chi^i$ with $\xi_i \geq 0$ then:
\begin{equation}
 \begin{array}{l}
  K_{\bullet m} ( {\bf x}, {\bf x}', \ldots, {\bf x}'''' ) = k ( \llp {\bf x}, {\bf x}', \ldots, {\bf x}'''' \ggp_m )
\end{array}
\end{equation}
is an $m$-dot-product kernel, and:
\begin{equation}
 \begin{array}{l}
  K_{\odot m} ( {\bf x}, {\bf x}', \ldots, {\bf x}'''' ) = \prod_i k ( x_i x_i' \ldots x_i'''' )
 \end{array}
\end{equation}
the $m$-direct-product kernel \citep{Sal2}.  A sample of $m$-kernels is 
presented in table \ref{tab:mkernel_eg}.  It is not difficult to see that the 
$m$-dot-product and $m$-direct-product kernels are free kernels with unweighted 
feature map $\vartheta_{\bf i} ({\bf x}) = x_0^{i_0} x_1^{i_1} \ldots 
x_{n-1}^{i_{n-1}}$ (${\bf i} \in \infset{N}^n$ is a multi-index) and weights:
\[
 \begin{array}{rl}
  \tau_{\bullet {\bf i}} \!\!\!&= \sqrt{ \frac{{\rm sum} \left( {\bf i} \right)!}{i_0! i_1! \ldots i_{n-1}!} \xi_{{\rm sum} \left( {\bf i} \right)} } \\
  \tau_{\odot {\bf i}}   \!\!\!&= \sqrt{ \xi_{i_0} \xi_{i_1} \ldots \xi_{i_{n-1}} } \\
 \end{array} 
\]
respectively.  It follows that all of the $m$-kernels in table 
\ref{tab:mkernel_eg} are free kernels, where we note that the SE $m$-kernel 
extension given in the table has unweighted feature map ${\mbox{\boldmath 
$\vartheta$}} ({\bf x}) = \| {\mbox{\boldmath $\tau$}}_e \odot {\mbox{\boldmath 
$\varphi$}}_e ({\bf x})\|_2^{-1} {\mbox{\boldmath $\vartheta$}}_e ({\bf x})$ 
and feature weights ${\mbox{\boldmath $\tau$}} = {\mbox{\boldmath $\tau$}}_e$, 
where ${\mbox{\boldmath $\vartheta$}}_e$ and ${\mbox{\boldmath $\tau$}}_e$ are 
the unweighted feature map and feature weights of the exponential 
$m$-dot-product kernel.

\subsection{Kernel Methods and Representor Theory}

Support vector machines (SVM) (and kernel methods more generally) are a family 
of techniques based around the concepts of structural risk minimization and the 
kernel trick \cite{Cor1,Ste3}.  At its most basic, if $\finset{A} = \{ ( {\bf 
x}^{\finset{A}}_i, y^{\finset{A}}_i) \in \mathbb{X} \times \mathbb{R} \}$ is a 
training set then the aim is to find an model:\footnote{A bias term is often 
included here, so $g_{\finset{A}} ({\bf x}) = {\bf w}_{\finset{A} }^{\tsp} 
{\mbox{\boldmath $\varphi$}} ({\bf x}) + b$.  Typically this results in an 
additional constraint on the dual (e.g. ${\rm sum} ({\mbox{\boldmath 
$\alpha$}}^{\finset{A}}) = 0$ for the examples given).  Alternatively the bias 
may always be incorporated into $K$ if required - precisely ${\mbox{\boldmath 
$\varphi$}} ({\bf x}) \rightarrow [ {\mbox{\boldmath $\varphi$}} ({\bf x}) ; 1 
]$, so the bias $b$ is included in ${\bf w}_{\finset{A}} \rightarrow [ {\bf 
w}_{\finset{A}} ; b ]$ and the kernel is adjusted as $K ({\bf x}, {\bf x}') 
\rightarrow K ({\bf x}, {\bf x}') + 1$.}
 \vspace{-0.1cm}
\begin{equation}
 \begin{array}{l}
  g_{\finset{A}} \left( {\bf x} \right) = {\bf w}_{\finset{A}}^{\tsp} {\mbox{\boldmath $\varphi$}} \left( {\bf x} \right)
 \end{array}
 \label{eq:svm_primal}
\end{equation}
where ${\mbox{\boldmath $\varphi$}} : \mathbb{X} \to \mathbb{R}^d$ is implied 
by a Mercer kernel $K$ and the weights ${\bf w}_{\finset{A}} \in \mathbb{R}^d$ 
solve:
 \vspace{-0.1cm}
\begin{equation}
 \begin{array}{c}
  \!\!\!\!\mathop{\min}\limits_{{\bf w}_{\finset{A}} \in \infset{R}^d} R = r \left( \frac{1}{2} \left\| {\bf w}_{\finset{A}} \right\|_2^2 \right) + \frac{1}{\lambda} \sum_i E \left( y^{\finset{A}}_i, g_{\finset{A}} \left( {\bf x}^{\finset{A}}_i \right) \right)\!\!\!\!
 \vspace{-0.1cm}
 \end{array}
 \label{eq:theprimal_svm}
\end{equation}
where $r$ is strictly monotonically increasing, $E$ is an empirical 
risk function, and $\lambda$ controls the trade-off between empirical risk 
minimization and regularization.  By representor theory \cite{Ste3}: 
\vspace{-0.1cm}
\begin{equation}
 \begin{array}{l}
  \exists {\mbox{\boldmath $\alpha$}}^{\finset{A}} \in \infset{R}^{|\finset{A}|} \mbox{ s.t. }
  {\bf w}_{\finset{A}} = \sum_i \alpha^{\finset{A}}_i {\mbox{\boldmath $\varphi$}} \left( {\bf x}^{\finset{A}}_i \right)
 \end{array}
 \label{eq:svm_normrep}
\end{equation}
and hence $g_{\finset{A}} ({\bf x}) = \sum_i \alpha^{\finset{A}}_i K 
({\bf x}, {\bf x}^{\finset{A}}_i)$.  Note that the weight vector ${\bf 
w}_{\finset{A}}$ is not (explicitly) present in this expression, and may be 
likewise removed from the training problem - for example in ridge regression 
(LS-SVM \cite{Suy1}) we let $E(y,g) = \frac{1}{2} (y-g)^2$, $r(v) = v$, so 
(\ref{eq:theprimal_svm}) becomes:
\vspace{-0.1cm}
\begin{equation}
 \begin{array}{l}
  \mathop{\min}\limits_{{\mbox{\scriptsize\boldmath $\alpha$}}^{\finset{A}} \in \infset{R}^{\left| \finset{A} \right|}} \frac{1}{2} {\mbox{\boldmath $\alpha$}}^{\finset{A}} {}^{\tsp} \left( {\bf K}_{\finset{A}} + \lambda {\bf I} \right) {\mbox{\boldmath $\alpha$}}^{\finset{A}} - {\bf y}^{\finset{A}} {}^{\tsp} {\mbox{\boldmath $\alpha$}}^{\finset{A}} 
\vspace{-0.1cm}
 \end{array}
 \label{eq:svm_thedualregress}
\end{equation}
where ${\bf K}_{\finset{A}} \in \infset{R}^{|\finset{A}| \times |\finset{A}|}$, 
$K_{\finset{A}i,j} = K ({\bf x}^{\finset{A}}_i, {\bf x}^{\finset{A}}_j)$; 
and for binary classification $y^{\finset{A}}_i = \pm 1$ and we may choose 
$E(y,g) = \max \{ 0, 1-yg \}$ (hinge loss), $r(v) = v$, so 
(\ref{eq:theprimal_svm}) becomes:\vspace{-0.1cm}
\begin{equation}
 \begin{array}{l}
  \mathop{\min}\limits_{{\mbox{\scriptsize\boldmath $\alpha$}}^{\finset{A}} \in \infset{R}^{\left| \finset{A} \right|}, {\bf 0} \leq {\bf y}^{\finset{A}} \odot {\mbox{\scriptsize\boldmath $\alpha$}}^{\finset{A}} \leq \frac{1}{\lambda} {\bf 1}} \frac{1}{2} {\mbox{\boldmath $\alpha$}}^{\finset{A}} {}^{\tsp} {\bf K}_{\finset{A}} {\mbox{\boldmath $\alpha$}}^{\finset{A}} - {\bf 1}^{\tsp} \left| {\mbox{\boldmath $\alpha$}}^{\finset{A}} \right|
 \end{array}
 \label{eq:svm_thedualclass}
\end{equation}
where $|{\mbox{\boldmath $\alpha$}}^{\finset{A}}| \in 
\infset{R}^{|\finset{A}|}$ is the element-wise absolute of ${\mbox{\boldmath 
$\alpha$}}^{\finset{A}}$.

\section{Gaussian Processes and Weight Prior Tuning with Free Kernels} \label{sec:gpdesc}

A gaussian process is a distribution on a space of functions \cite{Mac3,Ras2}.  
The following introduction takes the weight-space perspective \cite{Ras2}, but 
rather than modeling $f$ as a linear function in the feature space defined by 
the feature map ${\mbox{\boldmath $\varphi$}} : \infset{X} \to \infset{R}$ 
implied by a given Mercer kernel $K : \infset{X} \times \infset{X} \to 
\infset{R}$, we instead model $f$ as a linear function in the {\em unweighted} 
feature space defined by the {\em unweighted} feature map ${\mbox{\boldmath 
$\vartheta$}} : \infset{X} \to \infset{R}$ implied by a given {\em free kernel} 
(see definition \ref{def:def_freekernel}) $K_m : \infset{X}^m \to \infset{R}$, 
with a weight prior ${\bf v} \sim \normdist ({\mbox{\boldmath $\mu$}}, {\mbox{\boldmath 
$\Sigma$}}_{\mbox{\boldmath $\tau$}})$ defined by the feature weights implied by 
$K_m$.  As we show, this is identical to the usual derivation from a 
function-space perspective, but in weight space we obtain the key insight that 
the feature weights ${\mbox{\boldmath $\tau$}} \in \infset{R}^d$ implied by the 
free kernel $K_m$ control the relative importance of the different features.  
Moreover, as we will show subsequently, it is straightforward to generate, in a 
principled manner, free kernels with the same unweighted feature map but 
distinct feature weights, allowing us to {\em tune} the weight-prior of our 
Gaussian process to better model $f$ and hence accelerate our Bayesian 
optimizer.

Let $K_m : \infset{X}^m \to \infset{R}$ be a (given) free kernel with implied 
(unweighted) feature map ${\mbox{\boldmath $\vartheta$}} : \infset{X} \to 
\infset{R}^d $ and feature weights ${\mbox{\boldmath $\tau$}} \in 
\infset{R}^d$.  We model $f$ using the unweighted feature map:
\begin{equation}
 \begin{array}{l}
  f \left( {\bf x} \right) = {\bf v}^{\tsp} {\mbox{\boldmath $\vartheta$}} \left( {\bf x} \right)
 \end{array}
 \label{eq:linmodel}
\end{equation}
assuming a weight prior ${\bf v} \sim \normdist ({\mbox{\boldmath $\mu$}}, 
{\mbox{\boldmath $\Sigma$}}_{\mbox{\boldmath $\tau$}})$, where 
${\mbox{\boldmath $\Sigma$}}_{\mbox{\boldmath $\tau$}} = {\rm diag} 
({\mbox{\boldmath $\tau$}}{}^{\odot 2})$.  Following the usual 
method\footnote{For example following \cite{Ras2}, the differences in the derivation are 
entirely cosmetic at this point, with ${\mbox{\boldmath $\vartheta$}}$ replacing 
${\mbox{\boldmath $\varphi$}}$, ${\bf v}$ replacing ${\bf w}$, and 
${\mbox{\boldmath $\Sigma$}}_{\mbox{\boldmath $\tau$}}$ replacing 
${\mbox{\boldmath $\Sigma$}}_p$.} we see that the posterior 
of ${\bf v}$ given observations $\finset{D} = \{({\bf x}_i, y_i) | y_i = f 
({\bf x}_i) + \epsilon_i, \epsilon_i \sim \normdist (0, \sigma^2) \}$ is ${\bf 
v} | \finset{D} \sim \normdist ({\bf m}_{{\bf v}|\finset{D}}, {\mbox{\boldmath 
$\Sigma$}}_{{\bf v}|\finset{D}})$:
\begin{equation}
 \begin{array}{rl}
   \!\!\!\!\!\!{\bf m}_{{\bf v}|\finset{D}}                     &\!\!\!\!= {\mbox{\boldmath $\mu$}} + {\mbox{\boldmath $\Sigma$}}_{{\mbox{\boldmath $\tau$}}\!} {\mbox{\boldmath $\Theta$}}_{\finset{D}} \left( {\mbox{\boldmath $\Theta$}}_{\finset{D}}^{\tsp} {\mbox{\boldmath $\Sigma$}}_{{\mbox{\boldmath $\tau$}}\!} {\mbox{\boldmath $\Theta$}}_{\finset{D}} + \sigma^2 {\bf I} \right)^{-1} {\bf y} \\
   \!\!\!\!\!\!{\mbox{\boldmath $\Sigma$}}_{{\bf v}|\finset{D}} &\!\!\!\!= {\mbox{\boldmath $\Sigma$}}_{{\mbox{\boldmath $\tau$}}\!} - {\mbox{\boldmath $\Sigma$}}_{{\mbox{\boldmath $\tau$}}\!} {\mbox{\boldmath $\Theta$}}_{\finset{D}} \left( {\mbox{\boldmath $\Theta$}}_{\finset{D}}^{\tsp} {\mbox{\boldmath $\Sigma$}}_{{\mbox{\boldmath $\tau$}}\!} {\mbox{\boldmath $\Theta$}}_{\finset{D}} + \sigma^2 {\bf I} \right)^{-1} {\mbox{\boldmath $\Theta$}}_{\finset{D}}^{\tsp} {\mbox{\boldmath $\Sigma$}}_{{\mbox{\boldmath $\tau$}}\!}\!\!\!\!\!\!\!\!\! \\
 \end{array}
 \label{eq:gp_weight}
\end{equation}
where $\Theta_{\finset{D} i,j} = \vartheta_i ({\bf x}_j)$.  Substituting 
(\ref{eq:gp_weight}) into (\ref{eq:linmodel}) we find $f ({\bf x}) | \finset{D} 
\sim \normdist ( m_{f|\finset{D}} ({\bf x}), \Sigma_{f|\finset{D}} ({\bf x}, 
{\bf x}'))$:
\begin{equation}
 \begin{array}{rl}
   m_{f|\finset{D}} \!\left( {\bf x} \right) \!\!&\!\!\!= \!\mu \left( {\bf x} \right) + {\bf k}_{\finset{D}}^{\tsp} \left( {\bf x} \right) \left( {\bf K}_{\finset{D}} + {\sigma}^2 {\bf I} \right)^{-1} {\bf y} \\
   \!\!\!\!\!\!\Sigma_{f|\finset{D}} \!\left( {\bf x}, {\bf x}' \right) \!\!&\!\!\!= \!K \left( {\bf x}, {\bf x}' \right) - {\bf k}_{\finset{D}}^{\tsp} \!\!\left( {\bf x} \right) \left( {\bf K}_{\finset{D}} \!+\! {\sigma}^2 {\bf I}  \right)^{-1} \!{\bf k}_{\finset{D}} \left( {\bf x}' \right) \!\!\!\!\!\!
 \end{array}
 \label{eq:gp_first_mod}
\end{equation}
where ${\bf y}, {\bf k}_{\finset{D}} ({\bf x}) \in \infset{R}^{|\finset{D}|}$, 
${\bf K}_{\finset{D}} \in \infset{R}^{|\finset{D}| \times |\finset{D}|}$, 
$\mu ({\bf x}) := {\mbox{\boldmath $\mu$}}^{\tsp} {\mbox{\boldmath $\vartheta$}} ({\bf x})$, 
$k_{\finset{D} i} ({\bf x}) = K_2 ({\bf x}, {\bf x}_i)$, and 
$K_{\finset{D} i,j} = K_2 ({\bf x}_i, {\bf x}_j)$.

It is worth noting that all of the $m$-dot product and $m$-direct-product free 
kernels share the same unweighted feature map.  When used in the above, then, 
all $m$-dot-product and $m$-direct-product free kernels model $f$ as a linear 
function in the same (unweighted) feature space but apply different weight 
priors ${\bf v} \sim \normdist ({\mbox{\boldmath $\mu$}}, {\mbox{\boldmath 
$\Sigma$}}_{\mbox{\boldmath $\tau$}})$.  Moreover as the following theorem 
demonstrates, it is possible to {\em overwrite} the feature weights 
${\mbox{\boldmath $\tau$}}$ for a given free-kernel, and hence apply arbitrary 
(diagonal) weight-priors on our GP.
\begin{th_kernel_weight}
 Let $K_m$ be a free kernel with unweighted feature map ${\mbox{\boldmath 
 $\vartheta$}}$ and feature weights ${\mbox{\boldmath $\tau$}}$, and let 
 $\finset{A}^* = \{ ({\bf x}^{\finset{A}}_i, \alpha^{\finset{A}}_i) \in 
 \infset{X} \times \infset{R} \}$.  Then the family of functions:
 \begin{equation}
  \begin{array}{l}
   \!\!\!\!\!\!\!\!\!K^{\finset{A}}_m \!\left( {\bf x}, {\bf x}'\!, \ldots \right) = \mathop{\sum}\limits_{i,j} \!\alpha^{\finset{A}}_i \alpha^{\finset{A}}_j K_{m+2} \!\left( {\bf x}^{\finset{A}}_i, {\bf x}^{\finset{A}}_j, {\bf x}, {\bf x}'\!, \ldots \right)\!\!\!\!\!\!\!\!\!
  \end{array}
  \label{eq:kern_rewight_eq}
 \end{equation}
 indexed by $m$ are a free kernel with unweighted feature map ${\mbox{\boldmath 
 $\vartheta$}}_{\finset{A}} = {\mbox{\boldmath $\vartheta$}}$, and feature 
 weights ${\mbox{\boldmath $\tau$}}_{\finset{A}} = {\mbox{\boldmath $\tau$}} 
 \odot \sum_i \alpha^{\finset{A}}_i {\mbox{\boldmath $\vartheta$}} ({\bf 
 x}^{\finset{A}}_i)$.
 \label{th:th_kernel_weight}
\end{th_kernel_weight}
{\bf Proof: $\;$}
Using definition \ref{def:def_freekernel}:
\[
 \begin{array}{l}
  K^{\finset{A}}_m ( {\bf x}, {\bf x}', \ldots ) 
  = {\sum}_{i,j} \alpha^{\finset{A}}_i \alpha^{\finset{A}}_j K_{m+2} ( {\bf x}^{\finset{A}}_i, {\bf x}^{\finset{A}}_j, {\bf x}, {\bf x}', \ldots ) \\
  = {\sum}_{i,j} \alpha^{\finset{A}}_i \alpha^{\finset{A}}_j \!\llp {\mbox{\boldmath $\tau$}}^{\epow 2}, {\mbox{\boldmath $\vartheta$}} ( {\bf x}^{\finset{A}}_i ), {\mbox{\boldmath $\vartheta$}} ( {\bf x}^{\finset{A}}_j ), {\mbox{\boldmath $\vartheta$}} ( {\bf x} ), {\mbox{\boldmath $\vartheta$}} ( {\bf x}' ), \ldots \ggp_{m+3} \\
  = \llp ( {\mbox{\boldmath $\tau$}} \odot \sum_i \alpha^{\finset{A}}_i {\mbox{\boldmath $\vartheta$}} ({\bf x}^{\finset{A}}_i) )^{\epow 2}, {\mbox{\boldmath $\vartheta$}} ({\bf x}), {\mbox{\boldmath $\vartheta$}} ({\bf x}'), \ldots \ggp_{m+1} \\
  = \llp {\mbox{\boldmath $\tau$}}_{\finset{A}}^{\epow 2}, {\mbox{\boldmath $\vartheta$}}_{\finset{A}} ({\bf x}), {\mbox{\boldmath $\vartheta$}}_{\finset{A}} ({\bf x}'), \ldots \ggp_{m+1} \hfill \square
 \end{array}
\]

The implication of this is that, starting with a free kernel $K_m$, we may 
train an SVM on our auxiliary dataset $\finset{A}$ and $K_2$ to obtain a 
model (in terms of unweighted feature map):
\[
 \begin{array}{l}
  g_{\finset{A}} \left( {\bf x} \right) = {\bf v}_{\finset{A}}^{\tsp} {\mbox{\boldmath $\vartheta$}} \left( {\bf x} \right)
 \end{array}
\]
which by representor theory is parameterized by ${\mbox{\boldmath $\alpha$}}^{\finset{A}} 
\in \infset{R}^{|\finset{A}|}$:
\begin{equation}
 \begin{array}{l}
  {\bf v}_{\finset{A}} = {\mbox{\boldmath $\tau$}} \odot \sum_i \alpha^{\finset{A}}_i {\mbox{\boldmath $\vartheta$}} \left( {\bf x}^{\finset{A}}_i \right)
 \end{array}
 \label{eq:w_in_theta}
\end{equation}
Each element $v_{\finset{A}i}$ of ${\bf v}_{\finset{A}}$ controls the relative 
influence of that feature on the model - that is, a belief regarding the 
relative importance of feature $i$, where a large value $v_{\finset{A}i}$ 
implies and important feature - and theorem \ref{th:th_kernel_weight} provides 
a mechanism to directly build this belief (prior) into a Gaussian process via 
the construction of Mercer kernel $K^{\finset{A}}_2$.  The following example 
illustrates the operation:

{\bf Example 1 (XOR features)} Let the auxiliary dataset $\finset{A}$ be the 
famous XOR training set given by:

\hspace{0.2cm}
\begin{minipage}[c]{0.22\textwidth}
\begin{tabular}{| c || c | c |}
\hline
$i$ & ${\bf x}_i^{\finset{A}}$ & $y_i^{\finset{A}}$ \\
\hline
$0$ & $[ -1, -1 ]$ & $-1$ \\
$1$ & $[ +1, -1 ]$ & $+1$ \\
$2$ & $[ -1, +1 ]$ & $+1$ \\
$3$ & $[ +1, +1 ]$ & $-1$ \\
\hline
\end{tabular}
\end{minipage} 
\begin{minipage}[c]{0.22\textwidth}
\includegraphics[width=\textwidth]{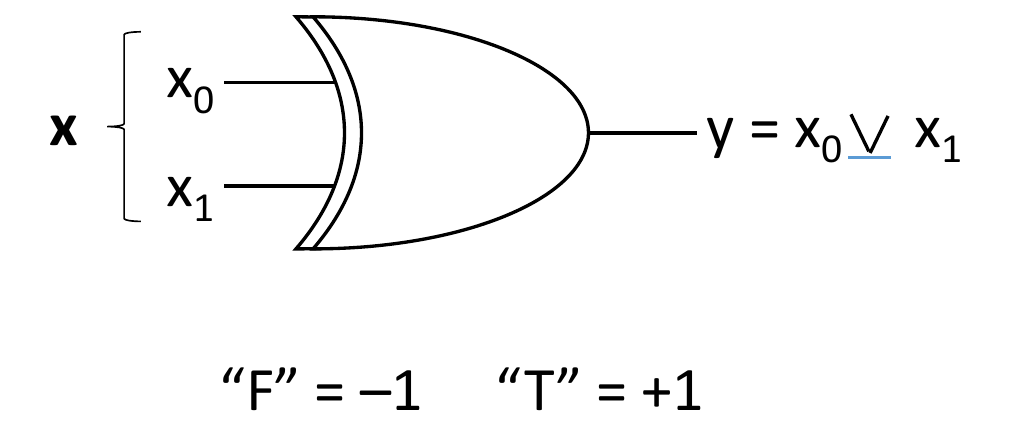}
\end{minipage} 


Let $K_m ({\bf x}, {\bf x}', \ldots) = (({\bf x}, {\bf x}', \ldots)_m+1)^2$ 
with ${\bf x}, {\bf x}', \ldots \in \infset{R}^2$ be the quadratic $m$-kernel, 
which may be readily seen to have an unweighted feature map ${\mbox{\boldmath 
$\vartheta$}} ({\bf x}) = [ 1, x_0, x_1, x_0^2, x_1^2, x_0x_1 ]$ and feature 
weights ${\mbox{\boldmath $\tau$}} = [ 1, \sqrt{2}, \sqrt{2}, 1, 1, \sqrt{2} 
]$.  Note that the only feature relevant to the XOR auxiliary training set is 
$x_0 x_1$, as $x_0 \veebar x_1 = x_0 x_1$ (recall $+1$ is ``T'' and $-1$ is 
``F'' in this representation).

If we train the SVM classifier (\ref{eq:svm_thedualclass}) with quadratic 
kernel $K_2$ and $C = 1$ on $\finset{A}$ we find ${\mbox{\boldmath 
$\alpha$}}^{\finset{A}} =[ -\frac{1}{8}, \frac{1}{8}, \frac{1}{8}, -\frac{1}{8} 
]$, and hence it is readily shown that the trained model (\ref{eq:svm_primal}) 
reduces to $g_{\finset{A}} ({\bf x}) = x_0 x_1$.  By theorem 
\ref{th:th_kernel_weight} (in particular (\ref{eq:kern_rewight_eq})) we may use 
$\finset{A}$ and ${\mbox{\boldmath $\alpha$}}^{\finset{A}}$ to construct the 
(modified) kernel:
\[
 \begin{array}{l}
  K^{\finset{A}}_2 \left( {\bf x}, {\bf x}' \right) = \sum_{i,j = 0,1,2,3} \alpha^{\finset{A}}_i \alpha^{\finset{A}}_j K_4 \left( {\bf x}^{\finset{A}}_i, {\bf x}^{\finset{A}}_j, {\bf x}, {\bf x}' \right) \\
  = \frac{1}{64} {\sum}_{i,j = 0,1,2,3} \left( 1 + {\bf x}^{\tsp} \left( {\bf x}^\finset{A}_i {\bf x}^{\finset{A}\tsp}_j \right) {\bf x}' \right)^2 
  = \frac{1}{2} x_0 x_1 x_0' x_1' \\
 \end{array}
\]
This constructed kernel has the same unweighted feature map ${\mbox{\boldmath 
$\vartheta$}}_{\finset{A}} ({\bf x}) = [ 1, x_0, x_1, x_0^2, x_1^2, x_0x_1 ]$ 
as the quadratic $m$-kernel $K_m$ from which it was derived, but with feature 
weights ${\mbox{\boldmath $\tau$}}_{\finset{A}} = [ 0, 0, 0, 0, 0, 
\frac{1}{\sqrt{2}} ]$ (as per theorem \ref{eq:kern_rewight_eq}).  This is 
reasonable in this case: as noted previously, $x_0x_1$ is the only relevant 
feature for the XOR training set, and the only feature with non-zero weight in 
the constructed kernel $K^{\finset{A}}_2$.

Used as a covariance for a GP, $K^{\finset{A}}_2$ applies a weight prior 
${\bf v} \sim \normdist ({\bf 0}, {\mbox{\boldmath $\Sigma$}}_{{\mbox{\boldmath 
$\tau$}}_{\finset{A}}})$, ${\mbox{\boldmath $\Sigma$}}_{{\mbox{\boldmath 
$\tau$}}_{\finset{A}}} = {\rm diag} ([ 0, 0, 0, 0, 0, \frac{1}{\sqrt{2}} ])$, 
which asserts a belief (derived from the auxiliary 
dataset $\finset{A}$) that only the weight corresponding to feature $x_0 x_1$ 
is important.\footnote{In most real-world examples we may expect that the 
feature weights will not be so sparse (as the SVM applies $2$-norm 
regularization on its weights rather than Lasso), and hence the belief asserted 
on the GP will be less dogmatic.}

\section{Accelerating Bayesian Optimization} \label{sec:apps}

Recall that our aim is to solve the problem:
\[
 \begin{array}{l}
  {\bf x} = {\rm argmax}_{{\bf x} \in \infset{X}} f \left( {\bf x} \right)
 \end{array}
\]
where $f$ is expensive to evaluate and we are given an auxiliary dataset 
$\finset{A} = \{ ({\bf x}^{\finset{A}}_i, y^{\finset{A}}_i) \in \infset{X} 
\times \infset{R} | y^{\finset{A}}_i = h(g({\bf x}^{\finset{A}}_i) + \nu_i), 
\nu_i \sim \normdist (0,\zeta^2) \}$ to accelerate the process (where 
$\finset{A}$ may include distilled knowledge e.g. from relevant patents, 
observations from related scenarios, human intuitions etc).  As noted 
previously, we assume that $f \sim \gp (0,\tilde{K})$ is a draw from a (zero 
mean) Gaussian process and that $g$ is well modeled using kernel $\tilde{K}$; 
and that $f$ and $g$ depend on similar {\em features} in the RKHS 
$\mathcal{H}_{\tilde{K}}$ induced by $\tilde{K}$.

As discussed in section \ref{sec:gpdesc}, for a free kernel $K_m$ we may 
extract weight priors from the auxiliary training set $\finset{A}$ by training 
an SVM, and then insert this as a prior in a Gaussian Process by constructing a 
kernel $K^{\finset{A}}_2$ using theorem \ref{th:th_kernel_weight}.  Taking as 
given that our assumptions regarding $f$ and $g$ are correct, the model $f \sim 
\gp (0,K^{\finset{A}}_2)$ will have more appropriate priors than a model $f \sim 
\gp (0,K)$ for a ``standard'' kernel $K$, as the weight prior reflects insight 
gained into the relative importance of the different features, and thus better 
modeling of $f$ using this derived kernel should improve the convergence rate 
of the Bayesian optimization of $f$.

Our algorithm is presented in algorithm \ref{alg:modded_bbo}.  As noted 
previously, algorithm \ref{alg:modded_bbo} does not assume direct knowledge of 
the covariance $K$ of the Gaussian process from which $f$ is drawn - rather, it 
infers (learns) $K$ from the auxiliary dataset $\finset{A}$ before proceeding.  
Unlike DT-MKL \cite{Dua2}, this search is not limited to the space spanned by a 
basis set of standard kernels - rather it is the space spanned by the 
unweighted feature map ${\mbox{\boldmath $\vartheta$}}$ of the $m$-kernel 
$K_m$.  For example if we use an exponential $m$-kernel then the search space 
is the space of all dot-product Mercer kernels.

\begin{algorithm}
\caption{Tuned-Prior Bayesian Optimization.}
\label{alg:modded_bbo}
\begin{algorithmic}
 \INPUT Auxiliary dataset $\finset{A}$, initial observations $\finset{D}_0$ of $f$.
 \INPUT Free kernel $K_m$ (for examples see table \ref{table:kern_table})
 \STATE Train SVM with dataset $\finset{A}$, kernel $K_2$ to obtain ${\mbox{\boldmath $\alpha$}}^\finset{A} \in \infset{R}^{|\finset{A}|}$.
 \STATE Construct re-weighted kernel $K^{\finset{A}}_2$:
\vspace{-0.2cm}
 \[
  \begin{array}{l}
   K^{\finset{A}}_2 ({\bf x}, {\bf x}') = {\sum}_{i,j} \alpha^{\finset{A}}_i \alpha^{\finset{A}}_j K_4 ( {\bf x}, {\bf x}', {\bf x}^{{\finset{A}}}_i, {\bf x}^{{\finset{A}}}_j )
  \end{array}
 \]
 \STATE \vspace{-0.2cm}Modeling $f \sim \gp ( 0, K^{\finset{A}}_2 )$, proceed:
 \FOR{$t=0,1,\ldots,T-1$}
 \STATE Select test point ${\bf x}_t = \argmax_{\bf x} a_t ({\bf x})$ (see (\ref{eq:acqfns})).
 \STATE Perform Experiment $y_t = f ( {\bf x}_t ) + \epsilon$, $\epsilon \sim \normdist (0,\nu)$.
 \STATE Update $\finset{D}_{t+1} := \finset{D}_t \cup \{ ({\bf x}_t,y_t) \}$.
 \ENDFOR
\end{algorithmic}
\end{algorithm} 

For the acquisition function $a_t$ we tested expected improvement 
(EI) \citep{Moc1,Jon1} and GP upper confidence bound (GP-UCB) \citep{Sri1}:
\begin{equation}
 \!\!\!\begin{array}{rl}
  a_t^{\!\rm EI}  \!\!&\!\!\!= \!( m_{\!f|\finset{D}_t} ( {\bf x} ) \!-\! y^+ ) \Phi ( Z_t ( {\bf x} ) ) \!+\! \sigma_{\!f|\finset{D}_t} ( {\bf x} ) \phi ( Z_t ( {\bf x} ) ) \\
  a_t^{\!\rm UCB} \!\!&\!\!\!= \!m_{\!f|\finset{D}_t} ( {\bf x} ) \!+\! \sqrt{\beta_t} \sigma_{\!f|\finset{D}_t} ( {\bf x} )
 \end{array}\!\!\!\!\!\!\!\!\!
 \label{eq:acqfns}
\end{equation}
where $Z_t ({\bf x}) = (m_{f|\finset{D}_t}({\bf x}) - y^+)/\sigma_{f| 
\finset{D}_t}({\bf x})$; $\sigma_{f|\finset{D}_t}^2 ({\bf x}) = 
\Sigma_{f|\finset{D}_t} ({\bf x},{\bf x})$; $\phi(\cdot)$ and $\Phi(\cdot)$ are 
the PDF and CDF functions for the normal distribution; $y^+ = \max_i \{y_i\}$; 
and $\beta_t$ are constants \cite{Sri1}.  We denote the variants of our 
algorithm using these acquisition functions, respectively, as TP-EI-BO and 
TP-GPUCB-BO.

Note that algorithm \ref{alg:modded_bbo} is divided into two distinct steps, 
specifically pre-training using the auxiliary dataset $\finset{A}$ to obtain 
the covariance $K^{\finset{A}}_2$, and Bayesian Optimization using this 
covariance to model $f \sim \gp (0,K^{\finset{A}}_2)$.  We have used standard 
BO here, but in general any variant of BO using this model could be 
substituted as required.

\vspace{-0.2cm}
\section{Experimental Results}
\vspace{-0.2cm}

\vspace{-0.1cm}
We now present a number of experiments applying algorithm \ref{alg:modded_bbo} 
using GP-UCB and EI acquisition functions  (TP-GPUCB-BO and TP-EI-BO), using a 
normalized SE free kernel with hyper-parameters (kernel length-scale and 
regularisation parameter $\lambda$) selected to minimize leave-one-out (LOO) 
error on the auxiliary dataset $\finset{A}$ during pre-training. We have 
compared our algorithm with Bayesian optimization using a standard 
squared-exponential covariance function (GPUCB and EI)\footnote{For GPUCB and 
EI the hyper-parameters were tuned to minimize LOO error on $\finset{D}_t$ at 
each iteration $t$.}, as well as standard transfer learning (envGPUCB and 
diffGPUCB \citep{Joy1,Shi21}\footnote{We also include envEI and diffEI, which 
are like envGPUCB and diffGPUCB but using the EI acquisition function.}), 
Bayesian optimization using a covariance function learned via Domain-Transfer 
Multi-Kernel Learning (DT-MKL-GPUCB and DT-MKL-EI) \cite{Dua2} using the kernel 
mixture:
\begin{equation}
 \begin{array}{l}
  K \left( {\bf x}, {\bf x}' \right) = {\sum}_{i = 0,1,2} v_i K_i \left( {\bf x}, {\bf x}' \right)
 \end{array}
 \label{eq:kernmix}
\end{equation}
where $K_0$ is an SE kernel, $K_1$ a Mat{\'e}rn 1/2 kernel, $K_2$ a Mat{\'e}rn 
3/2 kernel (all hyper-parameters in (\ref{eq:kernmix}) were selected to 
minimize LOO error on the auxiliary dataset $\finset{A}$ in pre-training), and 
Bayesian optimization using an ARD-SE kernel with hyperparameters tuned to 
minimize LOO error on the auxilliary dataset (ARD-GPUCB and ARD-EI).  All 
experiments were normalized to $f, h \circ g : 
[-1,1]^n \to [0,1]$.

\vspace{-0.1cm}
All experiments run with SVMHeavy v7 \cite{svmheavyv7} (code available at \url{https://github.com/apshsh/SVMHeavy}).


\begin{figure*}
 \centering 
  \includegraphics[width=0.5\textwidth]{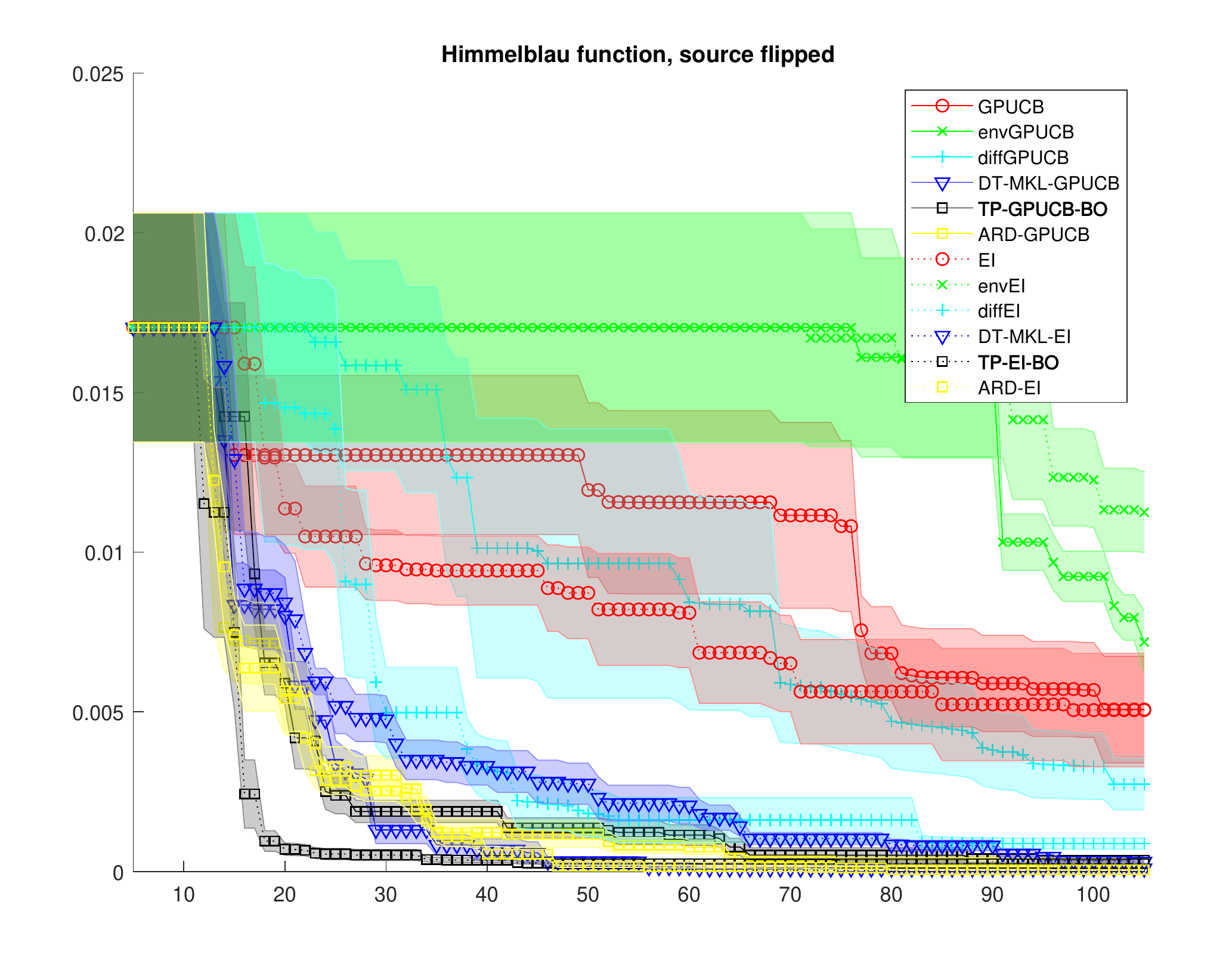} \hspace{-0.5cm}
  \includegraphics[width=0.5\textwidth]{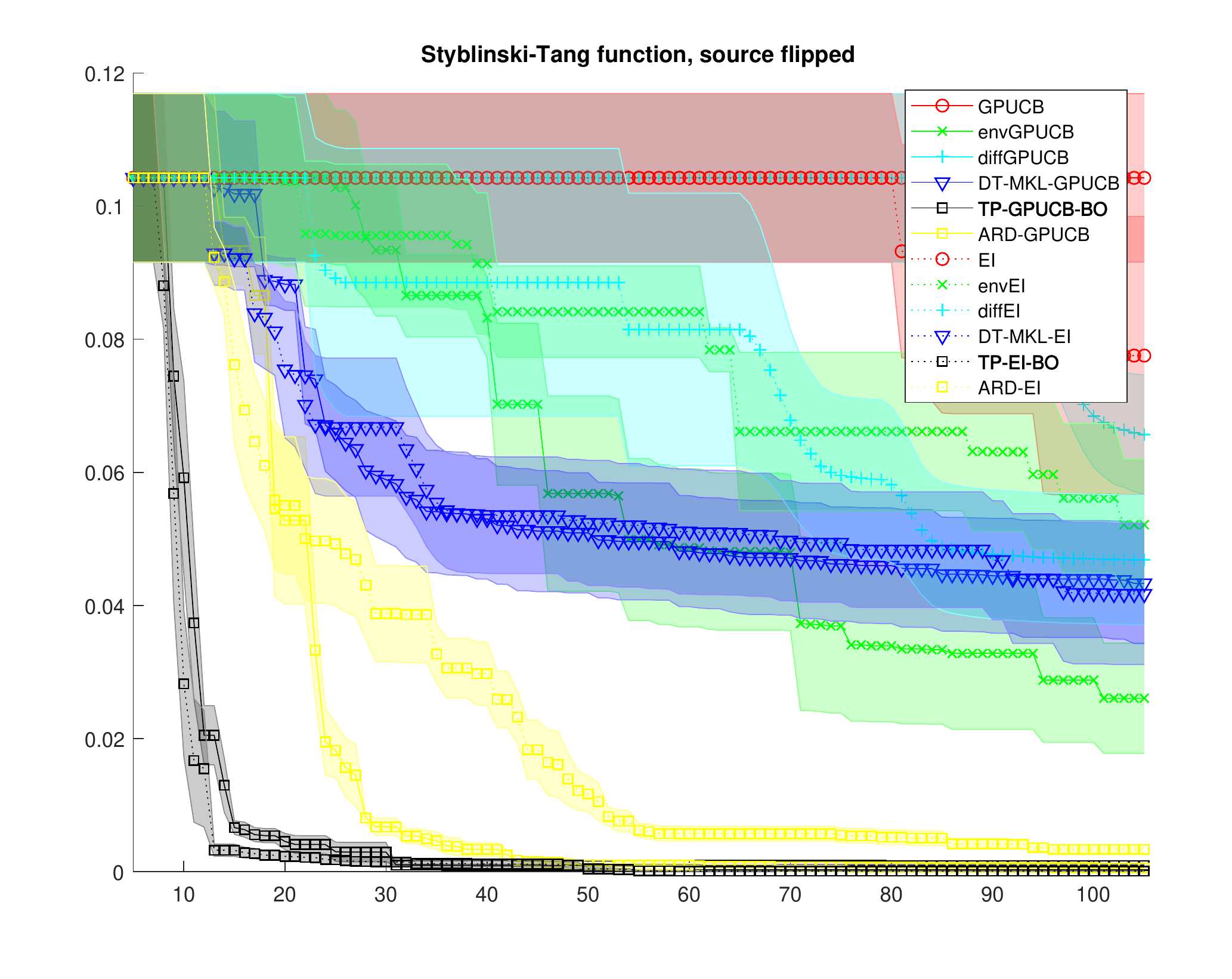} \\
  \vspace{-0.5cm}
  \includegraphics[width=0.5\textwidth]{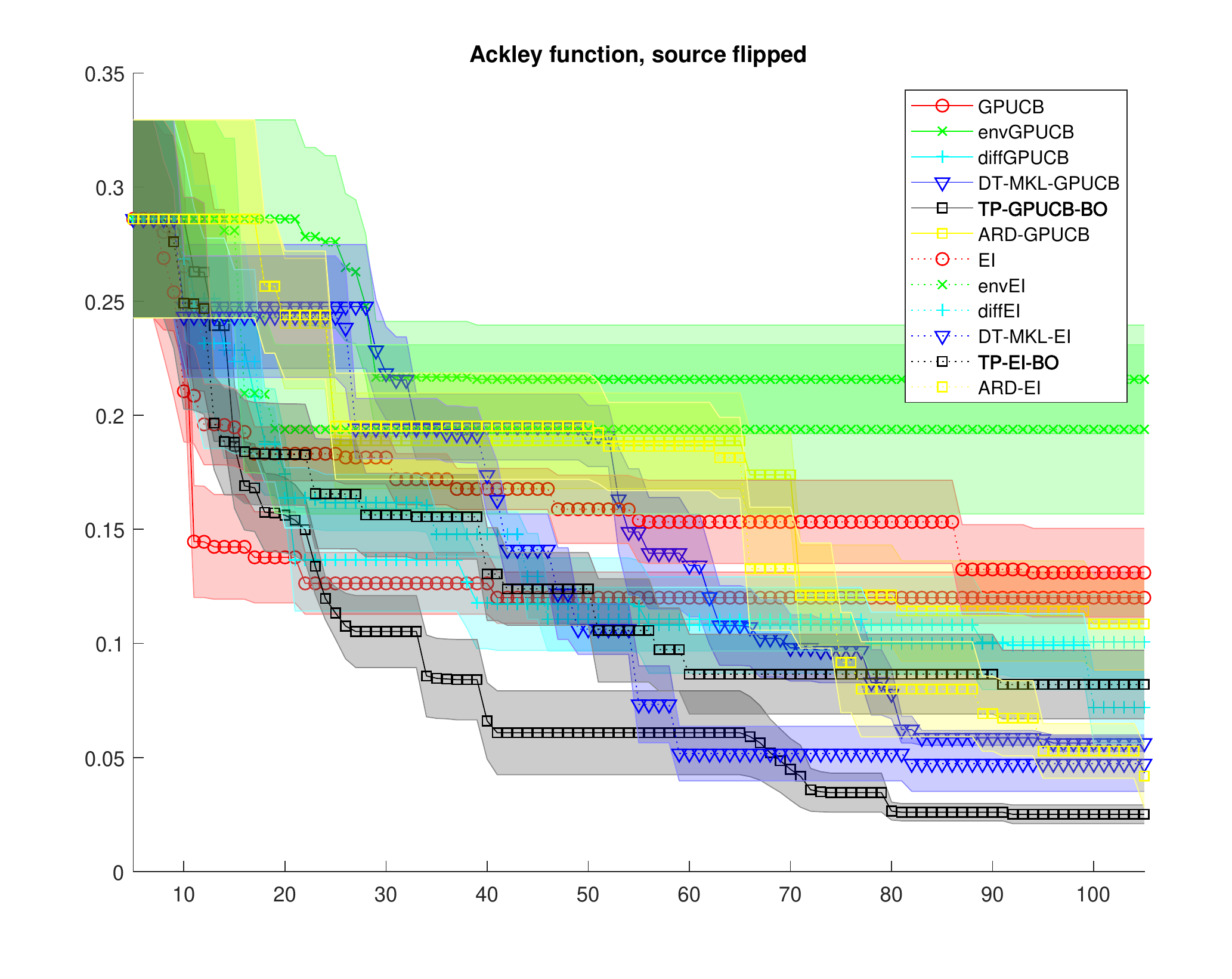} \hspace{-0.5cm}
  \includegraphics[width=0.5\textwidth]{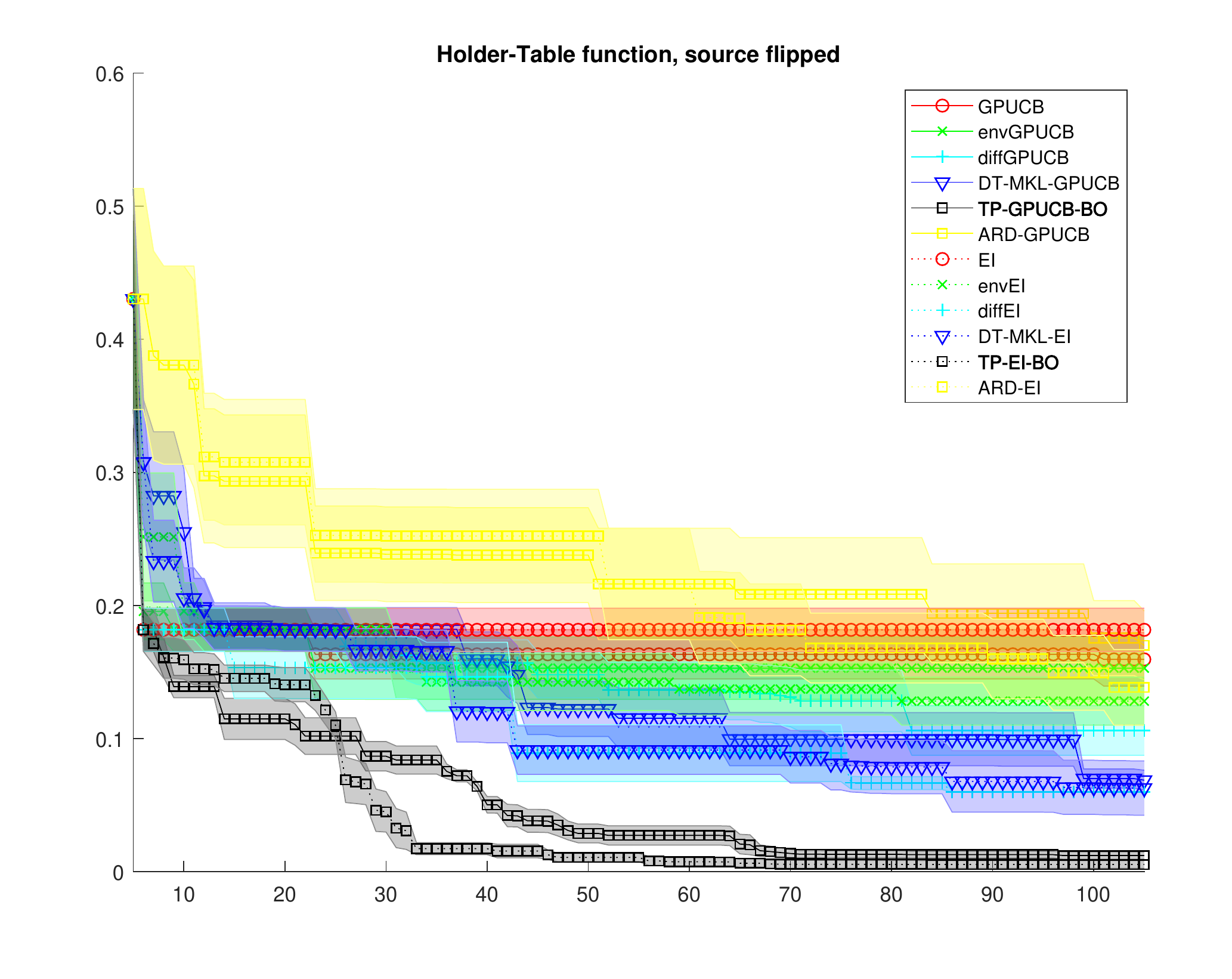} \\
  \vspace{-0.5cm}
  \includegraphics[width=0.5\textwidth]{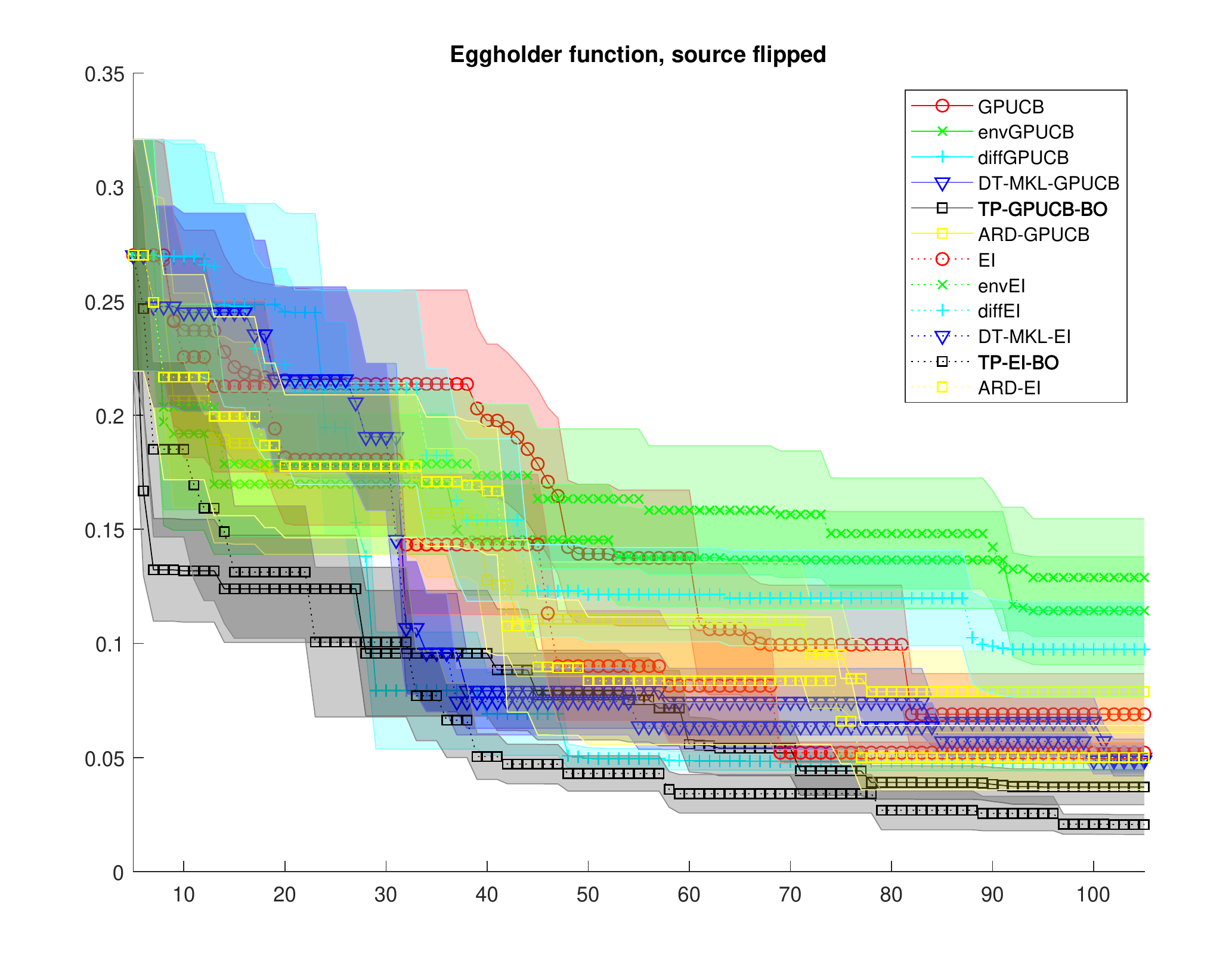} \hspace{-0.5cm}
  \includegraphics[width=0.5\textwidth]{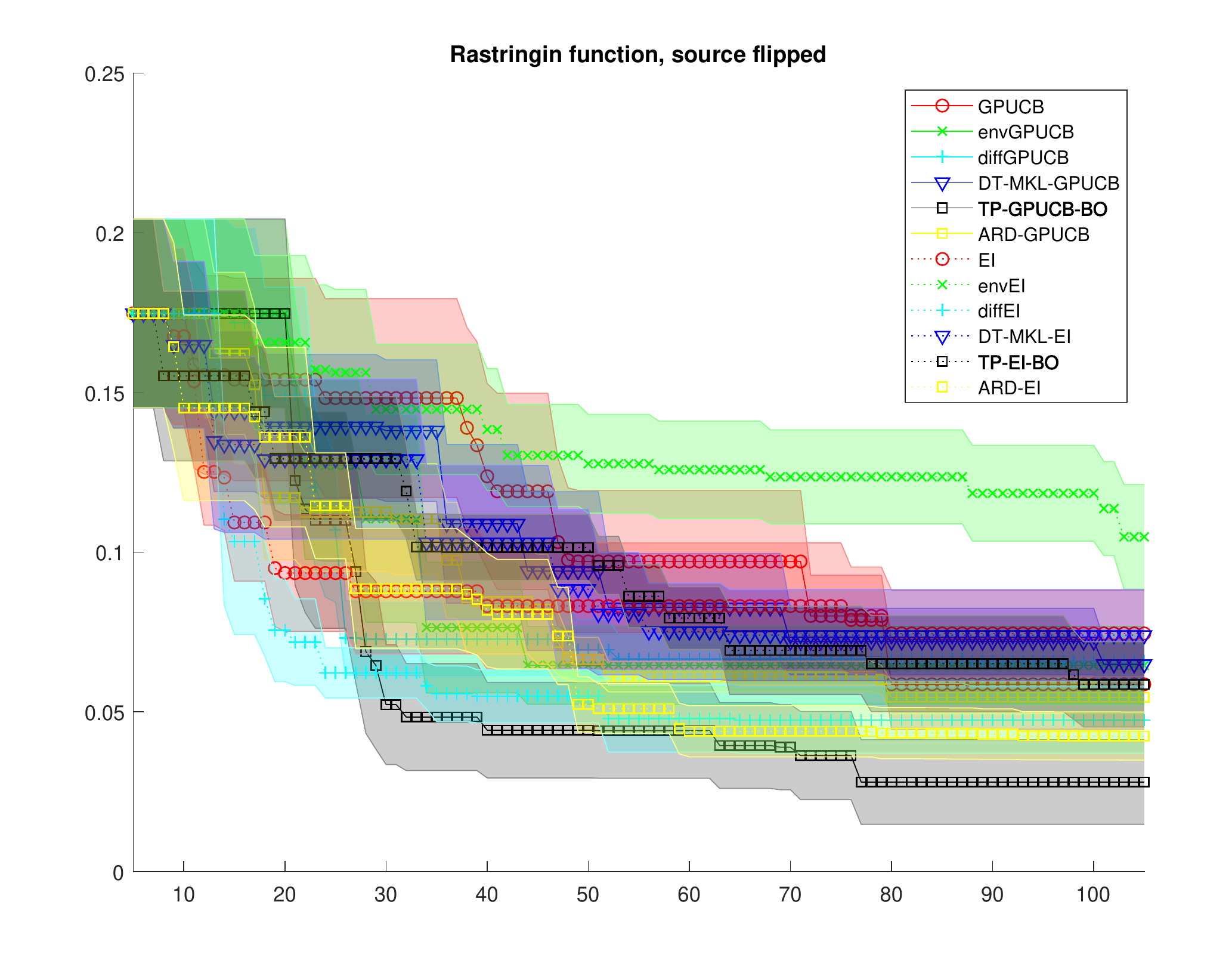} 
 \caption{Convergence for test functions with flipped auxiliary data $f = -h 
          \circ g$.  Variants of our method are shown in black.}
 \label{fig:eggares}
\end{figure*}

\vspace{-0.1cm}
\subsection{Experiment 1: Flipped Test Function}
\vspace{-0.2cm}

In this experiment we consider the optimization of the ($2$-dimensional) 
H{\"o}lder-Table, Himmelblau, Ackley, Styblinski-Tang, eggholder and Rastringin 
test functions.  In these experiments $50$ auxiliary datapoints were drawn 
from the flipped (negated) test function $h(g({\bf x})) = -f({\bf x})$ to make 
$\finset{A}$.  These were chosen for their non-trivial structure.

\begin{figure}
 \centering 
 \includegraphics[width=0.4\textwidth]{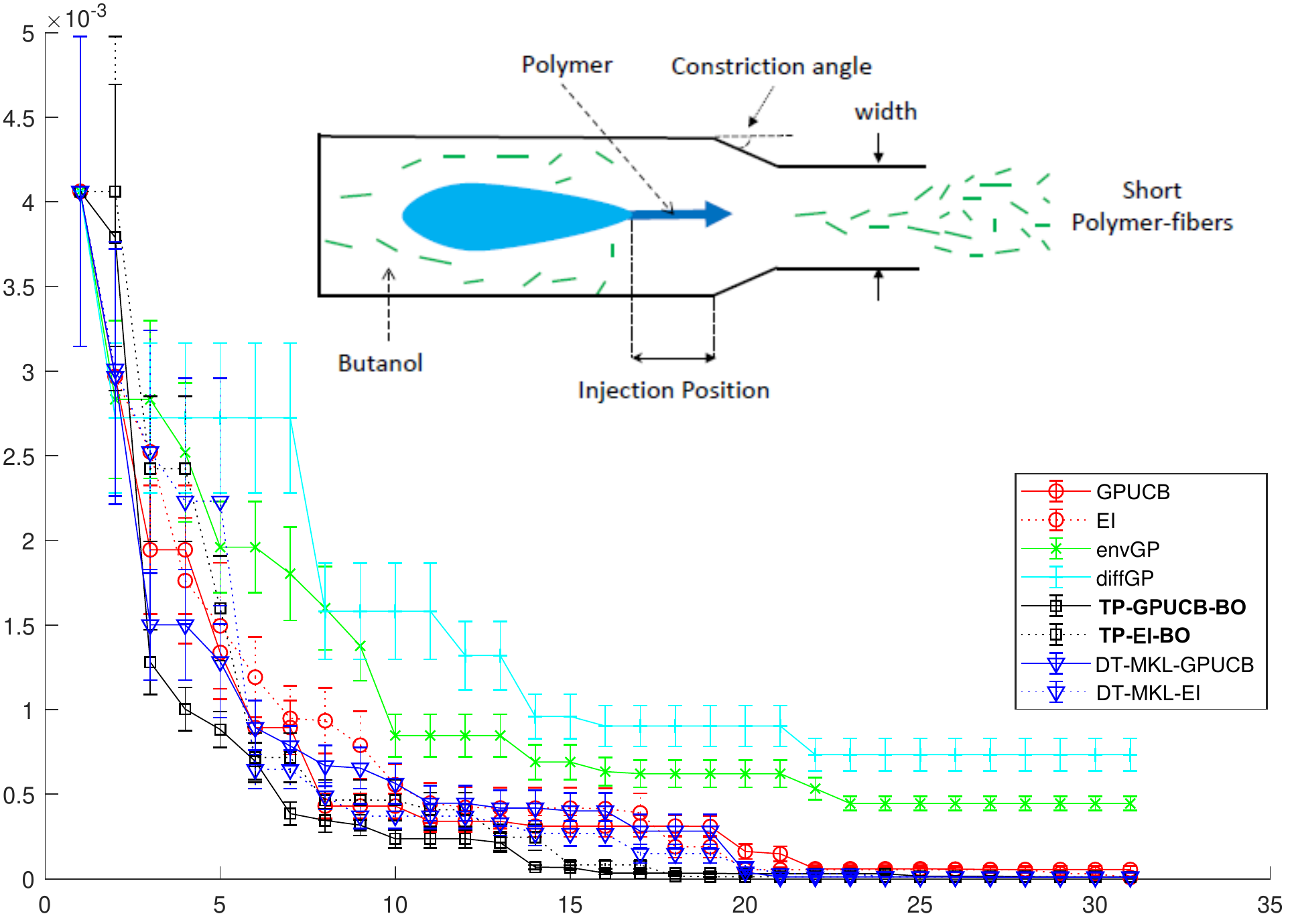} 
 \caption{Short Polymer Fibre design.  Comparison of algorithms in terms of 
          minimum squared distance from the set target versus iterations.  
          Device geometry for short polymer fibre injection shown as inset.}
 \label{fig:polyres}
 \label{fig:devgeom}
\end{figure}

Results are shown in figure \ref{fig:eggares}.  As expected in this situation, 
standard transfer learning algorithms (envGPUCB, diffGPUCB, envEI and diffEI) 
typically (though curiously not universally) perform worse than standard 
Bayesian optimization due to misdirecting auxiliary data; whereas DT-MKL helps 
convergence for all but the Rastringin function.  Our method performs well in 
all cases shown.  

\subsection{Experiment 2: Short Polymer Fibres}


In this experiment we have tested our algorithm on the real-world application 
of optimizing short polymer fibre (SPF) to achieve a given (median) target 
length \citep{Li6}.  This process involves the injection of one polymer into 
another in a special device \citep{Sut8}.  The process is controlled by $3$ 
geometric parameters (channel width (mm), constriction angle (degree), device 
position (mm)) and $2$ flow factors (butanol speed (ml/hr), polymer 
concentration (cm/s)) that parametrize the experiment (figure 
\ref{fig:devgeom}).  Two devices (A and B) were used.  Device A is armed by a 
gear pump and allows for three butanol speeds (86.42, 67.90 and 43.21), device 
B has a lobe pump and allows butonal speed 98, 63 and 48.  Our goal is to 
design a new short polymer fibre using Device B with median fibre length 
500$\mu$m, so:
\begin{itemize}
 \item Target: we aim to minimize $f({\bf x}) = (g_B({\bf x})-500)^2$, where 
       $g_B({\bf x})$ is the median fibre length ($\mu$m) produced by device B 
       with settings ${\bf x} \in \infset{R}^5$ as described above.  For 
       simulation purposes this is represented by a grid of $155$ experimental 
       observations from \citep{Li6}.
 \item Auxiliary dataset: $\finset{A}$ consists of $162$ experimental 
       observations of device A of the form $y^{\finset{A}}_i = (g_A({\bf x}^{\finset{A}}_i)-500)^2$, where 
       $g_A({\bf x})$ is the median fibre length produced by device A with 
       ${\bf x} \in \infset{R}^5$ \citep{Li6}.
\end{itemize}


Results are shown in figure \ref{fig:polyres}.  We note that, for this 
experiment, neither diffGP or envGP perform well, presumably due to the 
dissimilarities between $g_A$ and $g_B$ (but not their covariance structure).  
We note that our algorithm converges more quickly than the alternatives.

\vspace{-0.3cm}
\section{Notes and Future Directions}
\vspace{-0.3cm}

In the experiments we found that our method works well provided that the 
auxiliary data is sufficient to characterize the covariance of the 
optimization problem.  As counter-examples we found that test functions that 
vary significantly on small scales (eg. the Levi N.13 test function), and 
test functions with sparse, sharp features that are easily missed (eg. the 
Easom test function), our method did not improve convergence (see 
supplementary).  In the first case (Levi N.13) this is due to the inability to 
characterize such short-scale variation from the (comparatively) small dataset 
$\finset{A}$, the remedy for which is increasing the size of $\finset{A}$ (but 
see shortly).  In the second case (Easom) difficulties arise due to the fact 
that the target is mostly flat, so $\finset{A}$ is constant for all points and 
subsequently we find ${\mbox{\boldmath $\alpha$}}^{\finset{A}} = {\bf 0}$ (and 
hence $K_2^{\finset{A}} ({\bf x}, {\bf x}') = 0$ everywhere).  We note that the 
vanishing kernel is easily detected, and moreover that all methods considered 
failed to make headway on this objective.


%

With regard to practicality, the computational cost of evaluating 
$K^{\finset{A}}_2$ is $O(|\finset{A}|^2)$, so the cost of solving ${\bf x}_t = 
{\rm argmax}_{\bf x} a_t ({\bf x})$ is $O(|\finset{A}|^2 |\finset{D}_T|^2 + 
|\finset{D}_T|^3 + n (|\finset{A}|^2 |\finset{D}_T| + |\finset{D}_T|^2))$, 
where term 1 is the (one-off) cost of computing ${\bf K}_{\finset{D}}$, term 2 
is the (one-off) cost of inverting same (this is cached), and term 3 
is the cost of evaluating the GP $n$ times for global optimization (calculating 
${\bf k}_{\finset{D}} ({\bf x})$ and matrix multiplications).  We found this 
acceptable provided that $|\finset{A}|$ is not too large ($|\finset{A}| 
\lesssim 200$ appears reasonable), but for larger auxiliary datasets some form 
of sparsification (e.g. \cite{Dem2}) will presumably be required.

Finally it is desirable to investigate the impact of weight-prior 
tuning on the regret bounds associated with whichever acquisition function is 
selected.  It is unclear at present how to proceed in this direction, so this 
remains an open problem.

\vspace{-0.3cm}
\section{Conclusion}
\vspace{-0.3cm}

In this paper we presented an algorithm that can use auxiliary data to 
construct a GP covariance corresponding to a more appropriate weight prior for 
the objective function than is present in ``standard'' covariance functions.  
Using this, we have shown how BO may be accelerated by modeling the objective 
function using this covariance function.  We have demonstrated our algorithm on 
a practical applications (short polymer fibre manufacture) and a number of test 
functions.  Finally we have discussed the applicability of our algorithm and 
future research directions.


\subsubsection*{Acknowledgments}

This research was partially funded by the Australian Government through the 
Australian Research Council (ARC)​​. Prof Venkatesh is the recipient of an ARC 
Australian Laureate Fellowship (FL170100006).

\newpage
\bibliographystyle{apalike}
\bibliography{universal}

\end{document}